\documentclass[letterpaper, 10 pt, conference]{ieeeconf}  

\IEEEoverridecommandlockouts                              

\usepackage{graphics} 
\usepackage{times}
\usepackage{helvet}
\usepackage{courier}
\usepackage{graphicx}
\usepackage{xcolor} 
\usepackage{amsmath} 
\usepackage{amssymb}  
\usepackage[hyphens]{url}

\usepackage{algorithm}
\usepackage{algpseudocode}
\usepackage{booktabs}

\makeatletter
\let\NAT@parse\undefined
\makeatother
\usepackage[pdfusetitle]{hyperref}
\hypersetup{pdfborder = 0 0 0,
  colorlinks = true,
  urlcolor = blue,
  linkcolor = black,
  citecolor = black,
}

\title{\LARGE \bf
Generating Executable Action Plans with Environmentally-Aware Language Models
}

\author{Maitrey Gramopadhye$^{1}$ and Daniel Szafir$^{1}$
\thanks{$^{1}$University of North Carolina at Chapel Hill, United States}%
}

\begin{document}

\maketitle
\thispagestyle{empty}
\pagestyle{empty}

\begin{abstract}

Large Language Models (LLMs) trained using massive text datasets have recently shown promise in 
generating action plans for robotic agents from high level text queries. However, these models typically do not consider the robot's environment, resulting in generated plans that may not actually be executable, due to ambiguities in the planned actions or environmental constraints. In this paper, we propose an approach to generate environmentally-aware action plans that agents are better able to execute. 
Our approach involves integrating environmental objects and object relations as additional inputs into LLM action plan generation to provide the system with an awareness of its surroundings, resulting in plans where each generated action is mapped to objects present in the scene. We also design a novel scoring function that, along with generating the action steps and associating them with objects, helps the system disambiguate among object instances and take into account their states.
We evaluated our approach using the VirtualHome simulator and the ActivityPrograms knowledge base and found that action plans generated from our system had a 310\% improvement in executability and a 147\% improvement in correctness over prior work. The complete code and a demo of our method is publicly available at \url{https://github.com/hri-ironlab/scene_aware_language_planner}.

\end{abstract}

\section{INTRODUCTION}

Recent work in the natural language processing (NLP) and machine learning (ML) communities has made tremendous breakthroughs in several core aspects of computational linguistics and language modeling driven by advances in deep learning, data, hardware, and techniques. These advancements have led to the release of pretrained large (million and billion+ parameter) language models (LLMs) that have achieved the state-of-the-art across a variety of tasks such as text classification, generation, and summarization, question answering, and machine translation, that demonstrate some abilities to meaningfully understand the real world \cite{brown2020language, Devlin2019BERTPO, radford2019language, raffel2020exploring, Li2021ImplicitRO, Roberts2020HowMK, otter2020survey, torfi2020natural}. LLMs also demonstrate cross-domain and cross-modal generalizations, such as retrieving videos from text, visual question answering and task planning \cite{bommasani2021opportunities}. In particular, recent works have explored using LLMs to convert high-level natural language commands  to actionable steps (e.g. ``bring water'' $\rightarrow$ ``grab glass'', ``fill glass with water'', ``walk to table'', ``put glass on table'') for intelligent agents \cite{huang2022language, ahn2022can, suglia:embert, pashevich2021episodic, Sharma2022SkillIA, Li2022PreTrainedLM}. Trained on diverse and extensive data, LLMs have the distinct ability to form action plans for varied high-level tasks.

\begin{figure*}[tbp]
    \centering
    \includegraphics[width=0.9\textwidth]{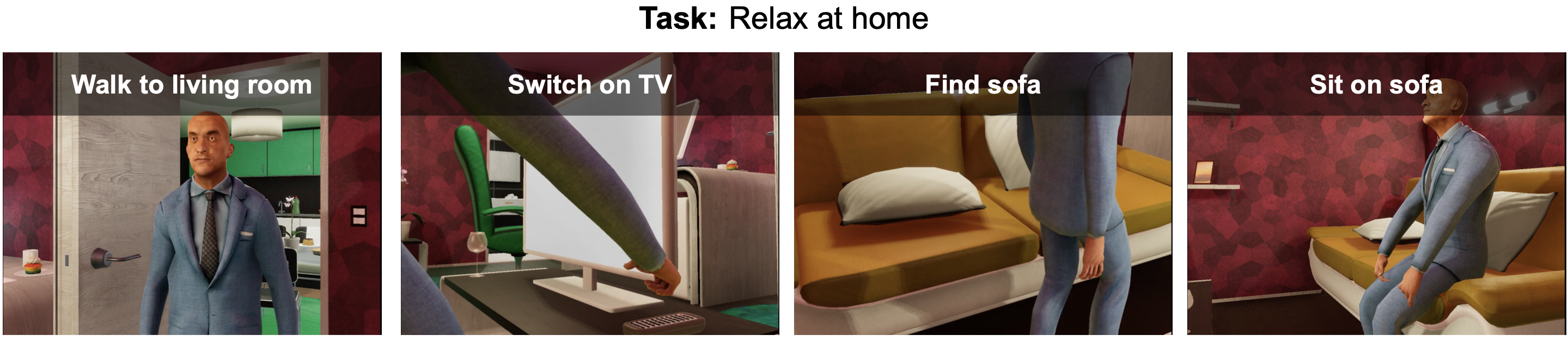}
    \caption{Visualization of an example action plan being executed in VirtualHome. Within the virtual home environment a simulated humanoid agent carries out the robot task sequences generated by our environmentally-aware language model.}
    \label{fig:img1}
\end{figure*}



While promising, the action steps generated by LLMs in prior work are not always executable by a robot platform. For instance, for a task ``clean the room'' a LLM might generate an output ``call cleaning agency on phone''; while being correct, this action plan might not be executable since the agent might not grasp the concept of ``call'' or have the ``phone'' object in it's environment. This limitation arises because LLMs are trained solely on large text corpora and have essentially never had any interaction with an embodied environment. As a result, the action steps they generate lack context on the robot's surroundings and capabilities. 

To address this issue, prior works have explored grounding LLMs by fine-tuning models using human interactions \cite{Li2022PreTrainedLM, Ouyang2022TrainingLM, reid2022wikipedia} or training models for downstream tasks using pretrained LLMs as frozen backbones \cite{Nair2021LearningLR, lynch2021language, blukis2021a, R3M, akakzia2021grounding, zellers-etal-2021-piglet, Hill2020HumanIW, pmlr-v162-humphreys22a}. However, these methods often require training on extensive annotated data, which can be expensive or infeasible to obtain, or can lead to loss of generalized knowledge from the LLM. Instead, recent research has investigated biasing LLM output without altering their weights by using prompt engineering \cite{huang2022language, ahn2022can, Wei2022ChainOT} or constraining LLM output to a corpus of available action steps defined \textit{a priori} that are known to be within a robot's capabilities \cite{huang2022language, ahn2022can}. This line of research focuses on methods that can utilise the capabilities of LLMs while preserving their generality and with substantially less additional annotated data.

While these systems effectively perform common sense grounding by extracting knowledge from an LLM, they employ a one-fits-all approach without considering the variations possible in the actionable environment. As a result, executing the action plans generated by these systems either requires approximations to the agent's environment or time-consuming and costly pretraining to generate an affordance score to determine the probability that an action will succeed or produce a favourable outcome towards task completion, given the current agent and environment states. Additionally, since prior systems are environment agnostic, it is not possible to use them to generate executable action plans for tasks requiring object disambiguation. 
For example, to generate correct action plans for tasks that require interaction with multiple objects with the same name
, the system needs to be able to distinguish among object instances. 

We propose a novel method to address these issues while generating low-level action plans from high-level task specifications. Our approach is an extension to Huang et al., 2022 \cite{huang2022language}. 
From an \textit{Example set} (see \S \ref{sec:eval}) using the ActivityPrograms knowledge base collected by Puig et al., 2018 \cite{puig2018virtualhome}, we sample an example similar to the query task and environment and use it to design a prompt for a LLM (details of which are given in \S \ref{section:prompt}). We then use the LLM to autoregressively generate candidates for each action step. To rank the generated candidates, we design multiple scores for the actions and their associated objects  
(see \S \ref{section:action} and \S \ref{section:object}). After the top candidate is selected, we append it to the action plan and repeat the process until the entire action plan is generated.

To evaluate our action plans, we use the recently released VirtualHome interface \cite{puig2018virtualhome} (Figure \ref{fig:img1} shows a visualization of an example action plan running in VirtualHome).  
We use several metrics (details in \S \ref{section:metrics}), including executability, Longest Common Sub-sequence (LCS), and final graph correctness to autonomously test generated action plans on VirtualHome. 
Overall, we found that our method increased action plan executability and correctness by 310\% and 147\% respectively over a state-of-the-art baseline.

\section{RELATED WORK}
Our work builds upon recent efforts in robotics to leverage the potential of LLMs. For instance, researchers are beginning to explore LLMs in the context of 
applying commonsense reasoning to natural language instructions \cite{chen2020enabling}, providing robotic agents with zero-shot action plans \cite{huang2022language}, and supplying high-level semantic knowledge about robot tasks \cite{ahn2022can}. Below, we review related research in task planning, LLMs, and action plan grounding.

\subsection{Task Planning}

The problem of task planning involves generating a series of steps to accomplish a goal in a constrained environment. Historically, this problem has been widely studied in robotics \cite{FIKES1971189, Sacerdoti1977ASF, Nau1999SHOPSH}, with most approaches solving it by optimizing the generated plan, given environment constraints, \cite{Toussaint2015LogicGeometricPA, ijcai2019p869} and using symbolic planning \cite{FIKES1971189, Nau1999SHOPSH}. Recently, machine learning methods have been employed to relax the constraints on the environment and allow higher-level task specifications by leveraging techniques such as reinforcement learning or graph learning to learn task hierarchy  \cite{NTP, xu2019rpn, NTG, NEURIPS2019_5c48ff18, savinov2018semiparametric, Ichter2021BroadlyExploringLT, Matuszek2012AJM, silver2022inventing, Garrett2020OnlineRI, Zhu2017VisualSP, Wu2021ExampleDrivenMR, Nair2020Hierarchical, xiali2020relmogen, li2019hrl4in, pu2020re, shah2022value}. However, most of these methods require extensive training from demonstrations, or explicitly encoded environmental knowledge and may not generalize to unseen environments and tasks. The use of LLMs, which encapsulate generalized world knowledge, may help plan for novel tasks and new environments.


\subsection{Large Language Models}
Large language models (LLMs) are language models, usually inspired by the transformer architecture \cite{NIPS2017_3f5ee243}, tens of gigabytes in size and trained on enormous amounts of unstructured text data. Recent advances in the field of natural language processing have shown that LLMs are useful for several downstream applications including interactive dialogue, essay generation, creating websites from text descriptions, automatic code completion, etc. \cite{brown2020language, Devlin2019BERTPO, raffel2020exploring, radford2019language}. During their pretraining, LLMs can accumulate diverse and extensive knowledge \cite{davison-etal-2019-commonsense, LPAQA, Petroni2019LanguageMA} that enables their use in applications beyond NLP, such as retrieving visual features \cite{Ilharco2020ProbingTM} and solving mathematical problems \cite{Cobbe2021TrainingVT, shen2021generate} or as pretrained models for other modalities \cite{lu2021fpt, tsimpoukelli2021multimodal}. In robotics, knowledge embedded in LLMs can be utilised to generate actionable plans for agents from high-level queries. However, in order for a plan to be executable by a robot, the outputs from the LLMs need to be grounded in the context of the robot's environment and capabilities. 

\subsection{Grounding Natural Language in Action Plans}

There has been considerable work towards grounding natural language in actionable steps. Prior research has focused on parsing natural language or analysing it as series of lexical tokens to remove ambiguity and map language commands to admissible actions \cite{artzi-zettlemoyer-2013-weakly, Misra2015EnvironmentDrivenLI, Misra2016TellMD, tenorth10webinstructions}. However, these methods usually require extensive, manually coded rules and thus fail to generalize to novel environments and tasks. More relevant to our approach, recent work has explored grounding language models using additional environment elements \cite{Sun2019VideoBERTAJ, li2019visualbert, Lu2019ViLBERTPT, NEURIPS2021_c6d4eb15, CLIP}. Techniques include prompting \cite{huang2022language, Wei2022ChainOT} and constraining language model outputs to admissible actions \cite{ahn2022can, suglia:embert, pashevich2021episodic, Sharma2022SkillIA}. To also ground the output of language models in the environment of the agent, prior works have tried using LLMs as fixed backbones, \cite{Nair2021LearningLR, blukis2021a, R3M, akakzia2021grounding, zellers-etal-2021-piglet, Hill2020HumanIW, pmlr-v162-humphreys22a, Lynch2020GroundingLI} fine-tuning or ranking model outputs through interactions with the environment \cite{Li2022PreTrainedLM, Ouyang2022TrainingLM, reid2022wikipedia}. Our work extends such approaches, where we use additional inputs from the environment (i.e., objects and their properties) to condition the model output without any fine-tuning of the LLM or  extra training to learn value functions for ranking LLM outputs. 

\section{APPROACH}
\label{section:approach}

\begin{figure}[tbp]
\centering
\includegraphics[width=.95\linewidth]{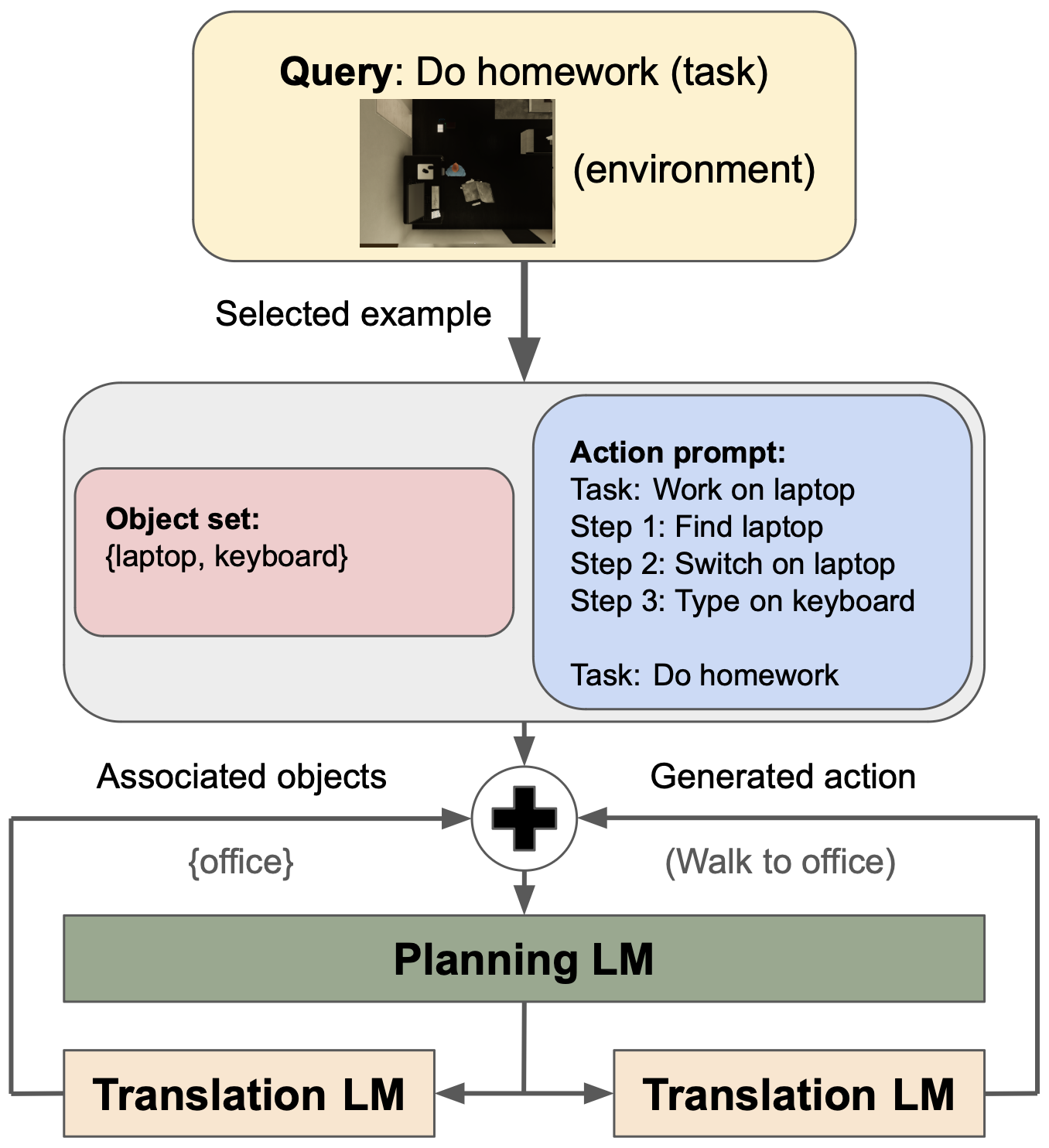}
\caption{An overview of our approach. We generate action plans by first selecting an example that has a similar task and environment to the query. We use this example to autoregressively prompt the Planning LM to generate an action plan and map the output to admissible actions and objects using the Translation LM.}
\label{fig:system}
\end{figure}

In this section, we discuss our proposed method to generate directly executable action plans from high-level tasks (Figure \ref{fig:system} provides a visual overview). Motivated by Huang et al., 2022 \cite{huang2022language}, our approach 
uses two language models, a \textit{planning} LM ($LM_P$) to generate the action plan and calculate a score for the similarity of an object with the other objects associated with the action plan; and a \textit{translation} LM ($LM_T$) to calculate embeddings for objects and actions.

\subsection{LLM Action Plan Prompt Generation}
\label{section:prompt}

Large language models have the ability to learn from context during inference, i.e., when autoregressively sampled, LLMs can generate meaningful text to complete or extend a given textual prompt \cite{brown2020language}. We leverage this capability in designing prompts for LLM sampling that generate action plans. Specifically, we select an example from an \textit{Example set} of task and action plans synthesized from the ActivityPrograms dataset (see \S \ref{sec:eval}) and construct a prompt for the LLM by prepending the example task and action plan to the current task.

We dynamically select the example during inference to design a prompt similar to the query. As in Huang et al., 2022 \cite{huang2022language}, we use the query task to select the example. However, one of our novel extensions is to also use the environment associated with the query to construct a prompt, keeping in mind the objects (and their states) the agent can currently interact with. For a query ($Q$) with task ``Play video games'', an example ($Ex_1$) with task ``play board games'' may be chosen considering just the task similarity. However, another example ($Ex_2$) with task ``Use the computer'' may be more relevant because the action plan for both $Q$ and $Ex_2$ would have actions involving similar objects, such as ``switch on computer'', ``type on keyboard'', ``push mouse'', etc., which may not be present in the action plan for $Ex_1$. 
Considering the environment in selecting the example may also help in disambiguating between examples with high task similarity but different objects. For example, a ``clean room'' action plan, which uses a rag, and a ``clean floor'' plan, which uses a mop, may both have high task similarity to a ``clean the house'' query task. Considering the objects present in the environment ($E$) of the query ($Q$) (e.g., a rag is present, but not a mop) can help determine the better example. 

We start by selecting $\{Q^i_e\}_{i=1}^{N_e}$ examples that have tasks $\{T^i_e\}_{i=1}^{N_e}$ similar to the task $T$ of the query $Q$. Here $N_e$ is a hyperparameter. We use the cosine similarity ($C$) of task embeddings to calculate task similarity given by:
\begin{equation*} \label{eg_task}
S_M(T,T_e) = C(LM_T(T), LM_T(T_e))
\end{equation*}
We then compare the environments $\{E^i_e\}_{i=1}^{N_e}$ of the selected examples with the environment ($E$) of the query ($Q$). An environment from a sample in our dataset is structured like a graph, with the graph nodes representing the available objects. The nodes also have information about object properties (eg. grabbable, openable, movable, etc.) and the current states of the objects (eg. clean, closed, etc.). The edges in the graph represent the relations between objects (eg. inside, on, facing, close to, etc.). We calculate the environment similarity as the mean of the intersection over unions of the nodes and edges respectively:
\begin{multline*} \label{eg_env}
S_G(E, E_e) = \frac{1}{2} \cdot (IoU(nodes(E), nodes(E_e)) + \\
IoU(edges(E), edges(E_e)))
\end{multline*}
Finally, from the selected $N_e$ examples, we select one example $Q^*_e$ that maximises the example score given by:
\begin{equation*} \label{eg_final}
S_M(T^*_e,T) + W_s \cdot S_G(E^*_e,E)
\end{equation*}
where $W_s$ is a hyperparameter. With the example task $T^*_e$, action plan $A^*_e$, and query task $T$, we form a prompt ($Pr_a = T^*_e + A^*_e + T$) for generating the action plan and a set of objects ($Pr_o$) associated with the actions plan $A^*_e$. We use $Pr_o$ to calculate the similarity scores between any new objects and the objects already associated with the action plan (See \S \ref{section:object}).

\subsection{Action Step Generation}
\label{section:action}

As in Huang et al., 2022 \cite{huang2022language}, we sample the $LM_P$ multiple times using prompt $Pr_a$ to get $k$ samples for each action step, and the LLM generation probability associated with each sample step ($P_a$). $P_a$ gives a score for how relevant the planner LM thinks the sample is to the current action plan and prompt. However, since the output of the language model is unconstrained, it can include infeasible steps that the agent cannot actually execute. To make sure the actions generated are executable, we map each sample to its closest admissible action step that maximises the action matching score given by:
\begin{equation*} \label{action_match}
P_{aM} = max(S_M(a_s, a_v); \forall a_v \in A_v)
\end{equation*}
Where $S_M(a_s, a_v) = C(LM_T(a_s), LM_T(a_v))$. $A_v$ is the corpus of all admissible atomic action steps, constructed by matching every possible action with every known object.

\subsection{Object Association}
\label{section:object}
Each action step discussed so far is of the format \texttt{[Action] <Object names>}. For an agent to execute an action step, it needs to associate the object names in the step with objects in the environment. However, action plans generated without considering the environment often contain object names that cannot be directly mapped to objects present in the environment, making the plan not executable.


We propose a way to autonomously associate objects in the query environment with action steps during action plan generation, using only a list of the objects present in the environment and their locations, without the use of any hard-coded information from the environment. As a result, our action plans can be executed directly in any query environment. Additionally, since our action plans consider the agent's environment, we are able to generate action plans for tasks that were previously infeasible. For example, a task ``set the table'' has an action plan with steps ``find a plate'', ``put plate on table'', ``grab cup'', ``put cup on table'', etc. repeated multiple times, each time for different ``plate'' and ``cup'' objects. A system that does not have any scene information may generate an action plan that has these action steps repeated in a correct order, but map all ``cup'' objects to the same ``cup'' and all ``plate'' objects to the same ``plate'' during execution, creating an incorrect final result. Since our method can distinguish among the ``cup'' and ``plate'' objects in the scene, we can associate different objects of the same name with repeated action steps, leading to correct execution.

Since the admissible action steps for our agent all follow the same schema, we first parse the step to extract the object names (e.g., ``put cup on table'' has the object names ``cup'' and ``table''). For each extracted object name ($\hat{o}$), 
we then assign an object from the environment. To do so, we first use a translation LM ($LM_T$) to 
find $o'$, the closest available object name to $\hat{o}$ that is present in the agent's environment, chosen by maximizing an object matching score given by:
\begin{equation*} \label{obj_match}
P_{oM} = max(S_M(\hat{o}, o); \forall o \in O)
\end{equation*}

\noindent where $O$ is the corpus of all object names present in the agent's current environment. We also calculate an object relevance score that denotes the similarity of $o'$ with the object names in the action plan of the example ($A^*$) and those associated with the previous steps of the generated action plan. For each object name $o^*$ in $Pr_o$, we construct the text ``$o^*$ and $o'$ are related''. We forward the text through the planning LM ($LM_P$) to calculate the associated cross-entropy loss ($L_{ce}$). We compute the object relevance score ($P_o$) as the mean of $-log(L_{ce})$ over all objects in $Pr_o$.

The object matching and relevance scores enable a consideration of environment objects for action plan generation, but will be the same for all objects in the scene with the same name. To disambiguate among such objects, we calculate an object disambiguation score ($P_{oD}$) as being inversely related to mean distance ($d_o$) of the object from the objects that the agent interacted with in the last generated step:
\begin{equation*} \label{obj_disamb}
P_{oD} = exp(\frac{-d_o}{100})
\end{equation*}

This score prioritizes objects that are near the agent's location in the prior step. Thus, promoting shorter and more efficient routes for the agent to complete an action step. For some object names, the location is absent in the dataset, and object disambiguation is impossible. In such cases we set $P_{oD}=0$. However since disambiguation occurs among objects with the same object name, this doesn't create an unfair bias for objects that have location information present. For each object, we also look for repetitions, i.e., instances where the current ($action$, $objects$) pair is found in the already generated action plan. For each such instance we incur a negative penalty to encourages the agent to interact with a different object with the same name. This negative score builds for every instance of repetition and can also reduce the overall score for the ($action$, $objects$) pair, deterring our action plans from loops of multiple $actions$ (e.g., ``walk to table'', ``walk to chair'' repeated over and over) that prior works suffer from. We assign the object with the greatest $P_{oD}$ to $\hat{o}$.




\subsection{Ranking and Termination}
\label{section:rank}

We take a weighted sum of the scores described above to rank all ($action$, $objects$) pairs:
\begin{equation*} \label{wt_sum}
W_a \cdot P_a + W_{aM} \cdot P_{aM} + W_o \cdot P_o + W_{oM} \cdot P_{oM} + W_{oD} \cdot P_{oD}
\end{equation*}

Where $W_a, W_{aM}, W_o, W_{oM}$ and $W_{oD}$ are hyperparameter weights for each score. If an action step has multiple objects, we calculate a mean for each of the 3 object scores ($P_{oM}, P_{o}, P_{oD}$), over the objects. We then select the highest ranking ($action$, $objects$) pair and append it to the action plan. We also append $action$ to the prompt for action generation ($Pr_a$) and $objects$ to the set for object relevance score ($Pr_o$). If the score for the highest ranking ($action$, $objects$) pair is below a \textit{cutoff} hyperparameter, we terminate the action plan.

The resulting action plans generated by the language model are in natural language. To execute and evaluate plans, we used the VirtualHome simulation platform (see \S \ref{sec:eval}), which required parsing this plan to create an action plan matching the VirtualHome agent schema. This parsing was performed using a predefined mapping since all natural language action steps follow a fixed pattern. 


\section{EVALUATION}
\label{sec:eval}

We evaluated our approach in generating environmentally-aware action plans against plans generated using the method in Huang et al., 2022 \cite{huang2022language}, the state-of-the-art baseline for a LLM action plan generation system, which does not consider environment information. We executed all plans using VirtualHome, a multi-agent simulation platform. VirtualHome provides diverse and customizable household environments that support a wide array of possible interactions in the form of atomic action steps. An atomic action step is specified by \texttt{[Action] <Objects> (object\_ids)} (e.g. 
\texttt{[PutBack] <glass> (2) <sink> (1)}).

For our experiments, we used the ActivityPrograms Knowledge Base released by Puig et al., 2018 \cite{puig2018virtualhome}. This dataset contains 292 unique high-level household tasks, with 1374
unique action plans and 6201 unique environments in total extracted from VirtualHome, and task and action plan samples manually annotated by Amazon Mechanical Turk workers. Each data point consists of a high-level task, a graphical representation of the agent's environment, and an action plan consisting of atomic actions. Out of the 292 tasks, 285 tasks have action plans and environments that execute without error in the VirtualHome interface. We performed our experiments on these 285 tasks, randomly split into 3 sets: an \textit{Example set} of 160 tasks, a \textit{Validation set} of 25 tasks and a \textit{Test set} of the remaining 100 tasks.

\subsection{Metrics} \label{section:metrics}

We evaluate our action plans across three metrics: executability, longest common sequence, and correctness.


\textbf{Executability} measures whether the generated actions follow a logical order and satisfy the constraints of the environment (e.g., action preconditions are met by preceding actions, action plan objects are present in the environment, and object states support planned actions). 
We executed the action plan step-by-step on the VirtualHome interface and calculated the number of steps that executed without the action plan failing. We report the percentage of steps that executed as the action plan executability score. 

Following Puig et al., 2018 \cite{puig2018virtualhome}, we computed \textbf{Longest Common Sub-sequence} (LCS) 
as the length of the longest common sub-sequence of steps between a generated action plan and a ``ground truth'' action plan from the ActivityPrograms dataset written by human annotators; divided by the length of the longer action plan. We required that the arrangement of the steps in the sub-sequence remains the same, but allowed gaps between them. LCS provides a metric to understand the number of correct actions being generated and also their short term order, while penalizing longer action plans that have irrelevant or repeated actions. However, it does not judge the correctness of the action plan as a whole because it does not consider the position of the longest common sequence in the action plan. 
Also, as LCS only compares the natural language action plans, it does not offer a way to judge whether the action plan is able to disambiguate among objects of the same name. 

We propose \textbf{Final Graph Correctness} as a new metric to evaluate the final correctness of the environment graph 
after execution of the action plans. We executed each generated action plan in VirtualHome up to the last executable step and extracted the graph for the resulting environment ($E_o$). We then compared this graph with the graph of the environment formed after executing the ground truth action plan ($E_G$). We computed the set of nodes and edges that changed in the initial environment ($E_{init}$) after executing the output and ground truth action plans respectively: $Nodes_o$, $Edges_o$, $Nodes_G$, $Edges_G$. The final graph correctness was calculated as the mean of intersection over unions of the two sets of nodes and edges thus obtained:
\begin{multline*}
    Final\ correctness = \frac{1}{2} \cdot (IoU(Nodes_o, Nodes_G) + \\
    IoU(Edges_o, Edges_G))
\end{multline*}

\subsection{Experimental Setup}

We evaluated and ablated over the \textit{Test set} (100 tasks) and used the \textit{Example set} (160 tasks) for prompt engineering. The \textit{Validation set} (25 tasks) was used for hyperparameter search. We ran all our experiments on Google Colab using a NVIDIA A100-SXM4-40GB GPU.

In action plans generated by the baseline, objects in action steps were not originally associated with the objects in the environment. For each object in each action step, we randomly selected an object of the same name if it is available, and assigned it to the action step.

We used open-source resources from Hugging Face Transformers \cite{Wolf2019HuggingFacesTS} and SentenceTransformers \cite{Reimers2019SentenceBERTSE} for our model choices. Our primary results used \texttt{GPT2-large} \cite{radford2019language} as the Planning LM and \texttt{all-roberta-large-v1}  \cite{Liu2019RoBERTaAR} as the Translation LM.

\begin{figure}[tbp]
\centering
\includegraphics[width=0.75\linewidth]{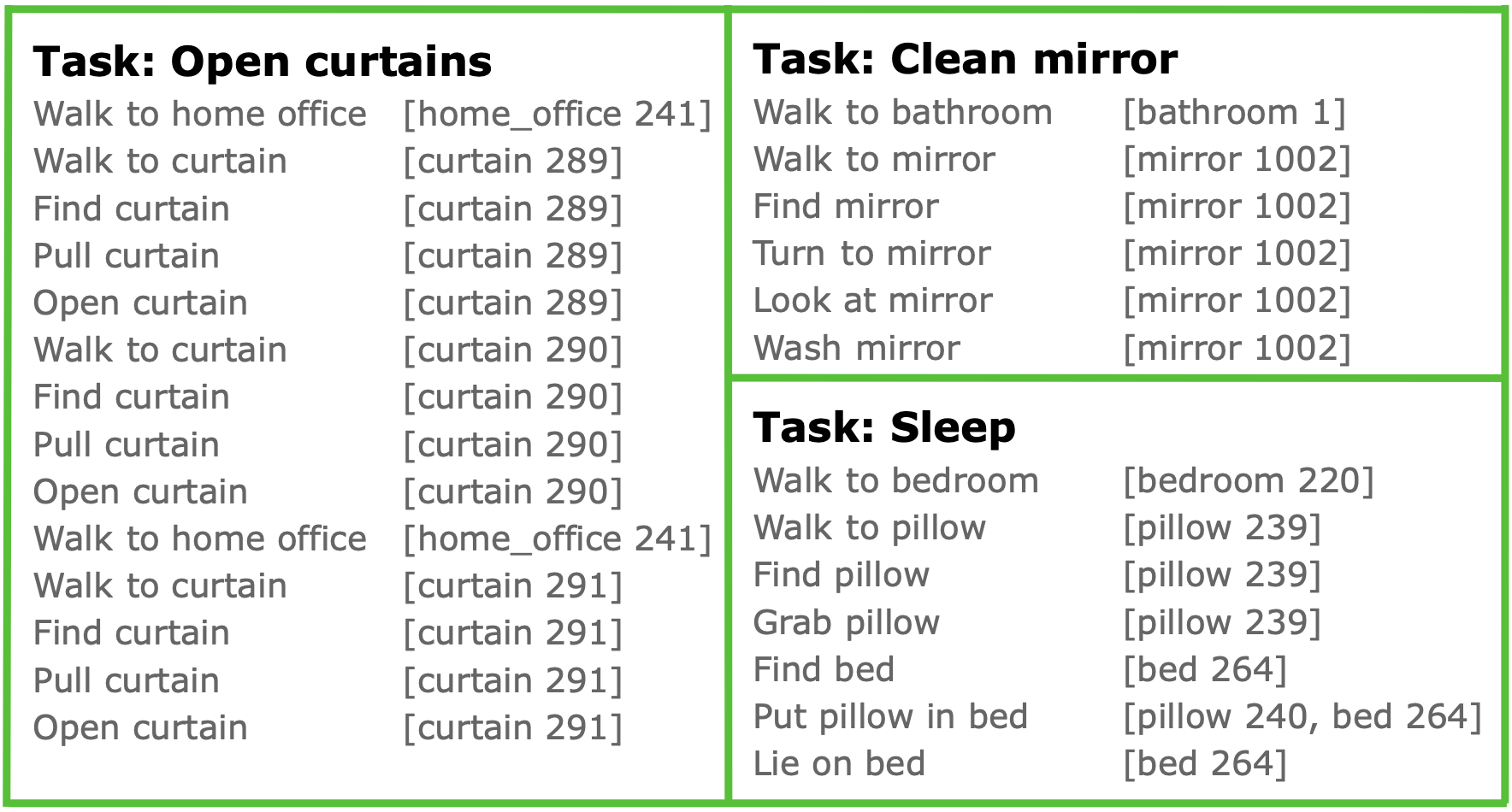}
\caption{Example plans generated by our system. For each action step, matched environment objects with ids are identified in brackets. Our system can handle plans containing actions with multiple objects (e.g., pillow and bed) and can consider multiple objects of the same name (e.g., curtain).}
\label{fig:examples}
\end{figure}

\section{RESULTS}

In this section we present our results. 
The step count is measured as the number of steps and all other metrics reported are a percentage of 100, unless mentioned otherwise. Some action plans generated by our method are illuminated in Figure \ref{fig:examples}.

\subsection{Environment Aware Action Plans}

We computed the action plans for each of the samples in our \textit{Test set}, using our method and the baseline \cite{huang2022language} that doesn't use any scene information. As each task can have multiple action plans and associated environments, we conducted 5 runs every time we generated an action plan; in each run we randomly selected an action plan and environment for each task. We report the average results over all the runs. Table \ref{main_res} reports the mean step count, executability, LCS and final graph correctness for both methods.

\begin{table}[h!]
\centering
\caption{Evaluation results of generated action plans}
\label{main_res}
\begin{tabular}{ c | c c c c c}
 Method & Steps & Executability & LCS & Correctness\\ 
 Huang et al. & 5.574 & 16.396\% & 8.795\% & 33.312\%\\
 {\bf Ours} & 8.380 & {\bf 50.826\%} & {\bf 10.941\%} & {\bf 48.990\%}
\end{tabular}
\end{table}

We observed that on average our method produced longer action plans, which allowed the agent to complete more complicated and longer-horizon tasks. Our method, using the information from the scene, generated action plans that were not only more executable but also led to final environment graphs that were closer to the desired results. As seen from the average LCS, using the environment information to select the example and inform action step generation also led to action plans that were closer to the ground truth action plans. However, we found that the extra computation associated with the agent's environment slowed down our method as compared to the baseline. On average, computing a step with our method took $1.435$ seconds, as compared to the $0.838$ seconds of the baseline.

These results are promising as our method, which 
implicitly considers object states and properties in our example selection module, could be readily integrated with a robot perception stack during real deployments. 

\subsection{Ablations}

\begin{table}[h!]
\centering
\caption{Ablations of implementation choices}
\label{ablate_score}
\begin{tabular}{ c | c c c c}
Method & Steps & Executability & LCS & Correctness\\ 
Baseline & 5.57 & 16.396\% & 8.795\% & 33.312\%\\
+ Object scores \\ ($P_{oM}, P_{o}$) & 10.87 & 43.892\% & 8.749\% & 47.715\%\\
+ Object disamb. \\ ($P_{oD}$) & 7.92 & 46.353\% & 9.390\% & 48.056\%\\
+ Env. similarity \\ ($S_G$) & 8.39 & {\bf 49.786\%} & {\bf 11.140\%} & {\bf 48.695\%}\\
\end{tabular}
\end{table}

\subsubsection{Ablation of Implementation Choices}

We ablated the scores we propose and show results over the \textit{test set} in Table \ref{ablate_score}. Compared to the baseline, we saw our biggest jump in executability and final correctness by incorporating object matching with the action steps and using the object scores ($P_{oM}, P_{o}$) to inform action step ranking. The additional scores also encouraged longer and more complicated action plans. Adding in the disambiguation score ($P_{oD}$) enabled the agent to distinguish between objects with the same name and allowed for more accurate action plans. It also discouraged redundant or repeated actions, thus resulting in shorter and more executable plans. Finally, since we used the objects in the environment to inform action step selection, also including this information in example selection ($S_G$) boosted results further as it made the resulting prompt more relevant to the current environment of the agent.

\begin{table}[h!]
\centering
\caption{Action plans with multiple objects of the same name}
\label{ablate_disamb}
\begin{tabular}{ c | c c c c}
Method & Steps & Executability & LCS & Correctness\\
w/o Object disamb. & 19.44 & 42.578\% & 7.686\% & 59.718\%\\
w/ Object disamb. & 8.96 & {\bf 58.570\%} & {\bf 12.006\%} & {\bf 61.462\%}\\
\end{tabular}
\end{table}

To exactly evaluate the usefulness of object disambiguation, we separately evaluated and compared the performance of the $23$ action plans which had a ($action$, $objects$) pair repeated in the action plan when object disambiguation was omitted. The results for this ablation are shown in Table \ref{ablate_disamb}.

\subsubsection{Ablation over Planning LMs}

We ablated over different sized Planning LMs (ranging from 117M parameters to 1.5B parameters in size) from two families of models \cite{radford2019language, zhang2022opt}, while fixing \texttt{all-roberta-large-v1} as the Translation LM (Table \ref{ablate_planning}). We observed that within a family of models, the medium sized models (\texttt{GPT2-large}, \texttt{OPT-350M}) gave the best results. We found that the small models (\texttt{GPT2}, \texttt{OPT-125M}) resulted in shorter action plans that were unable to capture the details required in action steps and ended up generating high-level instructions which were vague and not executable by the agent (e.g. a task ``work in office'' generated an action plan - ``Step 1: Go to office'', ``Step 2: work''). On the other hand, the large models (\texttt{GPT2-xl}, \texttt{OPT-1.3B}) often generated complicated samples that couldn't effectively be mapped to any available actions and thus resulted in action plans that were not relevant to the query task (e.g., a task ``Shampoo hair'' generated a sample ``grab a shampoo bottle and get in the shower'' which couldn't be mapped to any atomic action step).


\begin{table}[h!]
\centering
\caption{Ablations of Planning LM}
\label{ablate_planning}
\begin{tabular}{ c | c c c c}
 Planning LM & Steps & Executability & LCS & Correctness\\ 
 GPT2 & 1.04 & 8.667\% & 0.188\% & 33.441\%\\
 GPT2-large & 8.39 & 49.786\% & 11.140\% & 48.695\%\\
 GPT2-xl & 9.58 & 32.018\% & 8.641\% & 47.132\%\\
 OPT-125M & 4.52 & 29.183\% & 10.681\% & 41.346\%\\
 OPT-350M & 6.23 & 51.442\% & 13.72\% & 50.101\%\\
 OPT-1.3B & 8.27 & 26.991\% & 7.932\% & 43.652\%
\end{tabular}
\end{table}

\subsubsection{Ablation over Translation LMs}

We also explored using different size models of Sentence BERT and Sentence RoBERTa \cite{Devlin2019BERTPO, Reimers2019SentenceBERTSE, Liu2019RoBERTaAR} for the Translation LM, fixing \texttt{GPT2-large} as the Planning LM (Table \ref{ablate_translate}).  We found that larger translation LMs (\texttt{stsb-bert-large}, \texttt{stsb-roberta-large}, \texttt{all-roberta-large-v1}) created better performing and slightly shorter action plans compared to smaller model variants (\texttt{stsb-bert-base}, \texttt{stsb-roberta-base}). We speculate that larger models may create more meaningful embeddings for actions and objects and thus better guide the Planning LM to correct actions; however they are harsher towards planning LM outputs that don't effectively match any atomic action step and thus caused the action plans to terminate faster.

\begin{table}[h!]
\centering
\caption{Ablations of Translation LM}
\label{ablate_translate}
\begin{tabular}{ c | c c c c}
 Translation LM & Steps & Executability & LCS & Correctness\\ 
 stsb-bert-base & 9.76 & 40.555\% & 10.892\% & 47.393\%\\
 stsb-bert-large & 9.07 & 47.412\% & 12.602\% & 49.711\%\\
 stsb-roberta-base & 9.53 & 38.479\% & 9.306\% & 47.386\%\\
 stsb-roberta-large & 8.49 & 40.279\% & 9.871\% & 48.383\%\\
 all-roberta-large-v1 & 8.39 & 49.786\% & 11.140\% & 48.695\%
\end{tabular}
\end{table}






\section{FUTURE WORK \& CONCLUSIONS}

In this paper, we propose a method to condition large language models on the information contained in an agent's environment to generate environmentally-aware action plans from high-level tasks. We propose multiple scores to rank the outputs of LLMs and ground them in the agent's surroundings. We discuss the performance of our generated action plans for complex and diverse tasks on the VirtualHome interface. While our results demonstrate improved performance in terms of plan executability and correctness over the state-of-the-art baseline, there are several areas for further improvements. For example, our approach makes implicit use of various object properties and states when selecting examples for prompt generation, but cannot make use of this information directly during plan generation. Future research might explore how to further improve plan executability by addressing this limitation. In addition, future work is needed to validate our approach in a real-world robot deployment, beyond the VirtualHome simulator, where object information can be derived from a robot perception and affordance reasoning stack. We hope this work spurs further investigations into how robotics may leverage LLMs in dynamic environments.

\bibliographystyle{IEEEtran}
\bibliography{bibliography} 



\appendix

\subsection{Pseudo-Code of the Algorithm}

This section presents the pseudo-code for generating an action plan in Algorithm \ref{voo_algo}. A detailed discussion of the method and action step scores is given in \S \ref{section:approach}.

\begin{algorithm} 
\caption{Generating environment aware action plans}
\label{voo_algo}
\begin{algorithmic}[1]
\State \textbf{Legend}
\State $LM_P$ : Planning language model for text completion
\State $LM_T$ : Translation language model for text embedding
\State $\{(T^i_e, A^i_e, E^i_e)\}^{N_e}_{i=1}$ : Example set, where each sample consists of a task $T_e$, action plan $A_e$ and environment $E_e$
\State $A_v$ : The corpus of all available atomic actions
\State $C$ : Cosine similarity function
\State $ExtractObjects$ : Extract objects from an atomic action
\State $S_M(a_1, a_2)$ : Similarity function for object or action embeddings = $C(LM_T(a_1), LM_T(a_2))$
\State $S_G(E_1, E_2)$ : Similarity function for environment graphs
\State \textbf{Input:} Test task $T$ and the query environment $E$
\State \textbf{Output:} Action plan comprising of executable action steps of type - ($action$, $objects$)
\\\hrulefill
\State Get example ($T^*_e, A^*_e, E^*_e$) that maximises $S_M(T^*_e,T) + W_s \cdot S_G(E^*_e,E)$
\State Initialize prompt for actions ($Pr_a$) with ($T^*_e + A^*_e + T$) and prompt for objects ($Pr_o$) with objects in $E^*_e$
\For{$step < max\_steps$}
    \State Sample $LM_P$ $k$ times to obtain candidate actions
    \For{each $action$}
        \State $objs$ = $ExtractObjects(action)$
        \State \parbox[t]{195pt}{$P_{aM}$ = $max(S_M(action, a_v); \forall a_v \in A_v$)\strut}
        \State \parbox[t]{195pt}{$P_a$ = generation probability returned by $LM_P$\strut}
        \State \parbox[t]{195pt}{$P_{oM}$ = $avg(max(S_M(\hat{o}, o_v); \forall o_v \in O); \forall \hat{o} \in objs$)\strut}
        \State \parbox[t]{195pt}{$P_o$ calculated from $LM_P$ using prompt $Pr_o$\strut}
        \State \parbox[t]{195pt}{$d_o$ = mean distance of the object from the objects interacted with in the last step\strut}
        \State \parbox[t]{195pt}{$P_{oD}$ = $exp(-d_o/100)$ \strut}
        \State \parbox[t]{195pt}{Ranking score = $W_a \cdot P_a + W_{aM} \cdot P_{aM} + W_o \cdot P_o + W_{oM} \cdot P_{oM} + W_{oD} \cdot P_{oD}$ \strut}
        \State \parbox[t]{195pt}{Get ($action$, $objects$) pair with highest ranking and append it to output.\strut}
        \State \parbox[t]{195pt}{Append $action$ to $Pr_a$ and $objects$ to $Pr_o$\strut}
        \State \parbox[t]{195pt}{Terminate action plan if ranking score $< cutoff$\strut}
    \EndFor
\EndFor
\end{algorithmic}
\end{algorithm}

\subsection{Admissible Atomic Action Steps}\label{section:appendix_steps}

This section discusses the admissible atomic action steps that the agent can perform. This corpus was constructed by matching every possible action in the dataset with every known object in the VirtualHome simulator. Each of the possible 42 actions is in natural language and has an injective mapping to the format acceptable to the VirtualHome simulator, as given in Table \ref{appendix_actions}. Table \ref{appendix_objs} lists all of the 228 objects that can be present in a VirtualHome environment.

\begin{table}[h!]
\centering
\caption{All admissible actions}
\label{appendix_actions}
\begin{tabular}{l l}
 {\bf Natural Language} & {\bf VirtualHome acceptable format}\\
 Sleep & [SLEEP]\\
 Stand up & [STANDUP]\\
 Wake up & [WAKEUP]\\
 Close $<Obj>$ & [CLOSE] $<Obj>$\\
 Cut $<Obj>$ & [CUT] $<Obj>$\\
 Drink $<Obj>$ & [DRINK] $<Obj>$\\
 Drop $<Obj>$ & [DROP] $<Obj>$\\
 Eat $<Obj>$ & [EAT] $<Obj>$\\
 Find $<Obj>$ & [FIND] $<Obj>$\\
 Grab $<Obj>$ & [GRAB] $<Obj>$\\
 Greet $<Obj>$ & [GREET] $<Obj>$\\
 Lie on $<Obj>$ & [LIE] $<Obj>$\\
 Look at $<Obj>$ & [LOOKAT] $<Obj>$\\
 Move $<Obj>$ & [MOVE] $<Obj>$\\
 Open $<Obj>$ & [OPEN] $<Obj>$\\
 Plug in $<Obj>$ & [PLUGIN] $<Obj>$\\
 Plug out $<Obj>$ & [PLUGOUT] $<Obj>$\\
 Point at $<Obj>$ & [POINTAT] $<Obj>$\\
 Pull $<Obj>$ & [PULL] $<Obj>$\\
 Push $<Obj>$ & [PUSH] $<Obj>$\\
 Put back $<Obj>$ & [PUTOBJBACK] $<Obj>$\\
 Take off $<Obj>$ & [PUTOFF] $<Obj>$\\
 Put on $<Obj>$ & [PUTON] $<Obj>$\\
 Read $<Obj>$ & [READ] $<Obj>$\\
 Rinse $<Obj>$ & [RINSE] $<Obj>$\\
 Run to $<Obj>$ & [RUN] $<Obj>$\\
 Scrub $<Obj>$ & [SCRUB] $<Obj>$\\
 Sit on $<Obj>$ & [SIT] $<Obj>$\\
 Squeeze $<Obj>$ & [SQUEEZE] $<Obj>$\\
 Switch off $<Obj>$ & [SWITCHOFF] $<Obj>$\\
 Switch on $<Obj>$ & [SWITCHON] $<Obj>$\\
 Touch $<Obj>$ & [TOUCH] $<Obj>$\\
 Turn to $<Obj>$ & [TURNTO] $<Obj>$\\
 Type on $<Obj>$ & [TYPE] $<Obj>$\\
 Walk to $<Obj>$ & [WALK] $<Obj>$\\
 Wash $<Obj>$ & [WASH] $<Obj>$\\
 Watch $<Obj>$ & [WATCH] $<Obj>$\\
 Wipe $<Obj>$ & [WIPE] $<Obj>$\\
 Release $<Obj>$ & [RELEASE] $<Obj>$\\
 Pour $<Obj 1>$ into $<Obj 2>$ & [POUR] $<Obj 1>$ $<Obj 2>$\\
 Put $<Obj 1>$ on $<Obj 2>$ & [PUTBACK] $<Obj 1>$ $<Obj 2>$\\
 Put $<Obj 1>$ in $<Obj 2>$ & [PUTIN] $<Obj 1>$ $<Obj 2>$
\end{tabular}
\end{table}

\begin{table}[h!]
\centering
\caption{All possible objects}
\label{appendix_objs}
\begin{tabular}{l | l | l | l}
razor & conditioner & paper towel & homework\\
bowl & fork & toaster & food oatmeal\\
music stand & ironing board & light & tooth paste\\
dirt & toilet paper & chef knife & pen\\
button & hanger & food pizza & child\\
bedroom & shoes & juice & arms both\\
bookshelf & pantry & clothes jacket & picture\\
pot & sponge & mousepad & cup\\
mirror & printer & address book & kitchen\\
clothes socks & computer & cd player & knife\\
duster & food apple & trashcan & electric shaver\\
water glass & shower & alarm clock & laptop\\
sofa & remote control & comforter & cutting board\\
notes & table cloth & food turkey & notebook\\
nightstand & cleaning bottle & cloth napkin & document\\
detergent & fridge & centerpiece & food carrot\\
pasta & living room & glass & cookingpot\\
painting & bed & mop & oven\\
cellphone & sink & food sugar & feet both\\
desk & dishrack & hair & kitchen cabinet\\
fryingpan & bills & wall clock & dry pasta\\
home office & coffee maker & food snack & freezer\\
console & love seat & wine glass & textbook\\
paper & ground coffee & mug & dvd player\\
lighting & food dessert & spoon & food fish\\
cat & television & clothes skirt & chair\\
filing cabinet & faucet & receipt & crackers\\
milk & radio & keyboard & washing machine\\
brush & novel & keys & curtain\\
bathroom & garbage can & broom & groceries\\
toothbrush & teeth & food bread & pajamas\\
food vegetable & microwave & clothes hat & folder\\
magazine & food chicken & legs both & clothes pants\\
closet & wall & dresser & sauce pan\\
couch & spectacles & after shave & board game\\
face soap & dishwasher & water & man\\
drinking glass & placemat & cabinet & printing paper\\
stereo & slippers & coin & kitchen counter\\
shampoo & toy & table & face\\
purse & hands both & bookmark & lamp\\
food cheese & beer & soap & dish soap\\
kids bedroom & sheets & cupboard & food food\\
scissors & bathtub & mop bucket & drying rack\\
headset & mouse & toilet & clothes scarf\\
clothes shirt & clothes dress & towel & creditcard\\
carpet & rag & coffee cup & food egg\\
window & coffee table & floor & controller\\
coffee filter & iron & bag & woman\\
dog & book & drawing & facial cleanser\\
phone & newspaper & coffee pot & envelope\\
pillow & food noodles & floor lamp & telephone\\
entrance hall & food cereal & stove & vacuum cleaner\\
plate & coffee & fly & mail\\
napkin & lightswitch & shredder & blanket\\
\multicolumn{2}{c|}{dining room} & \multicolumn{2}{|c}{bathroom counter}\\
\multicolumn{2}{c|}{instrument piano} & \multicolumn{2}{|c}{electrical outlet}\\
\multicolumn{2}{c|}{toothbrush holder} & \multicolumn{2}{|c}{cleaning solution}\\
\multicolumn{2}{c|}{clothes underwear} & \multicolumn{2}{|c}{laundry detergent}\\
\multicolumn{2}{c|}{video game console} & \multicolumn{2}{|c}{basket for clothes}\\
\multicolumn{2}{c|}{video game controller} & \multicolumn{2}{|c}{bathroom cabinet}
\end{tabular}
\end{table}

\newpage

\subsection{Sample Action Plan}

This section gives an example of an action plan from the ActivityPrograms knowledge base. The action plan consists of atomic action steps in the format described in \S \ref{section:appendix_steps}. Since each natural language action can be mapped to a VirtualHome acceptable format, each action plan represented in natural language also has a unique action plan in VirtualHome acceptable format (see Table \ref{appendix_ap}).

\begin{table}[h!]
\centering
\caption{Sample Action Plan from ActivityPrograms knowledge base}
\label{appendix_ap}
\begin{tabular}{l | l}
Walk to kitchen & [WALK] $<$kitchen$>$ (1)\\
Walk to dish soap & [WALK] $<$dish soap$>$ (1)\\
Find dish soap & [FIND] $<$dish soap$>$ (1)\\
Grab dish soap & [GRAB] $<$dish soap$>$ (1)\\
Find plate & [FIND] $<$plate$>$ (1)\\
Put dish soap on plate & [PUTBACK] $<$dish soap$>$ (1) $<$plate$>$ (1)\\
Find plate & [FIND] $<$plate$>$ (1)\\
Grab plate & [GRAB] $<$plate$>$ (1)\\
Wash plate & [WASH] $<$plate$>$ (1)\\
Rinse plate & [RINSE] $<$plate$>$ (1)\\
Find dishrack & [FIND] $<$dishrack$>$ (1)\\
Put plate on dishrack & [PUTBACK] $<$plate$>$ (1) $<$dishrack$>$ (1)
\end{tabular}
\end{table}

\subsection{Hyperparameter Tuning}\label{section:appendix_hyperparam}

We used the Validation set of size 25 to conduct a grid search for the hyperparameters given in Table \ref{appendix_hyperparam}.

\begin{table}[h!]
\centering
\caption{Hyperparameters}
\label{appendix_hyperparam}
\begin{tabular}{l | l}
{\bf Hyperparameter name} & {\bf Values considered}\\
Weight for Object Matching Score ($W_{oM}$) & \{0, 0.3, 0.5, 0.7\}\\
Weight for Object Relevance Score ($W_{o}$) & \{0, 0.1, 0.25\}\\
Weight for Object Disambiguation Score ($W_{oD}$) & \{0, 0.1, 0.3, 0.5\}\\
Weight for Environment Similarity ($W_{s}$) & \{0, 0.25, 0.5\}\\
Temperature modulating $LM_P$ \\ sample probability ($T$) & \{0, 0.3, 0.5\}\\
Action step score cutoff & \{0.8, 1.0, 1.2, 1.4, 1.6\}\\
\end{tabular}
\end{table}

\subsection{Qualitative Results for Action Plans}

This section presents some more action plans generated by our method.

\begin{table}[h!]
\centering
\begin{tabular}{l l}
\multicolumn{2}{c}{\bf Task: Watch TV}\\
\midrule
Walk to living room & [dining\_room 1]\\
Walk to remote control & [remote\_control 1000]\\
Find remote control & [remote\_control 1000]\\
Grab remote control & [remote\_control 1000]\\
Walk to couch & [couch 192]\\
Sit on couch & [couch 192]\\
Touch remote control & [remote\_control 1000]\\
Find television & [television 216]\\
Switch on television & [television 216]\\
Turn to television & [television 216]\\
Watch television & [television 216]\\
\bottomrule
\end{tabular}
\end{table}

\begin{table}[h!]
\centering
\begin{tabular}{l l}
\multicolumn{2}{c}{\bf Task: Surf internet}\\
\midrule
Walk to home office & [home\_office 246]\\
Walk to desk & [desk 282]\\
Find chair & [chair 283]\\
Sit on chair & [chair 283]\\
Find computer & [computer 321]\\
Switch on computer & [computer 321]\\
Turn to home office & [home\_office 246]\\
Walk to computer & [computer 321]\\
Turn to computer & [computer 321]\\
Type on home office & [home\_office 246]\\
\bottomrule
\end{tabular}
\end{table}

\begin{table}[h!]
\centering
\begin{tabular}{l l}
\multicolumn{2}{c}{\bf Task: Take dishes out of dishwasher}\\
\midrule
Walk to kitchen & [oven 229]\\
Walk to dishwasher & [dishwasher 1000]\\
Find dishwasher & [dishwasher 1000]\\
Open dishwasher & [dishwasher 1000]\\
\bottomrule
\end{tabular}
\end{table}

\begin{table}[h!]
\centering
\begin{tabular}{l l}
\multicolumn{2}{c}{\bf Task: Sort laundry}\\
\midrule
Wash clothes underwear & [clothes\_underwear 2306]\\
Wash clothes socks & [clothes\_socks 2029]\\
\bottomrule
\end{tabular}
\end{table}

\begin{table}[h!]
\centering
\begin{tabular}{l l}
\multicolumn{2}{c}{\bf Task: Sleep}\\
\midrule
Walk to bedroom & [bedroom 220]\\
Walk to pillow & [pillow 239]\\
Find pillow & [pillow 239]\\
Grab pillow & [pillow 239]\\
Find bed & [bed 264]\\
Put pillow in bed & [pillow 240, bed 264]\\
Lie on bed & [bed 264]\\
\bottomrule
\end{tabular}
\end{table}

\begin{table}[h!]
\centering
\begin{tabular}{l l}
\multicolumn{2}{c}{\bf Task: Hang keys}\\
\midrule
Walk to entrance hall & [dining\_room 41]\\
Walk to purse & [purse 1001]\\
Find purse & [purse 1001]\\
Open purse & [purse 1001]\\
Find keys & [keys 1002]\\
Grab keys & [keys 1002]\\
Close purse & [purse 1001]\\
\bottomrule
\end{tabular}
\end{table}

\begin{table}[h!]
\centering
\begin{tabular}{l l}
\multicolumn{2}{c}{\bf Task: Watch movie}\\
\midrule
Walk to living room & [dining\_room 49]\\
Walk to remote control & [remote\_control 1000]\\
Find remote control & [remote\_control 1000]\\
Grab remote control & [remote\_control 1000]\\
Walk to couch & [couch 230]\\
Sit on couch & [couch 230]\\
Touch remote control & [remote\_control 1000]\\
\bottomrule
\end{tabular}
\end{table}

\begin{table}[h!]
\centering
\begin{tabular}{l l}
\multicolumn{2}{c}{\bf Task: Vacuum}\\
\midrule
Walk to living room & [dining\_room 41]\\
Walk to closet & [bathroom 1]\\
Open closet & [bathroom 1]\\
Find vacuum cleaner & [vacuum\_cleaner 1001]\\
Grab vacuum cleaner & [vacuum\_cleaner 1001]\\
Pull vacuum cleaner & [vacuum\_cleaner 1001]\\
Plug in vacuum cleaner & [vacuum\_cleaner 1001]\\
Switch on vacuum cleaner & [vacuum\_cleaner 1001]\\
Pull vacuum cleaner & [vacuum\_cleaner 2420]\\
Push vacuum cleaner & [vacuum\_cleaner 1001]\\
Pull vacuum cleaner & [vacuum\_cleaner 2461]\\
Push vacuum cleaner & [vacuum\_cleaner 2420]\\
Pull vacuum cleaner & [vacuum\_cleaner 2571]\\
Push vacuum cleaner & [vacuum\_cleaner 2461]\\
Switch off vacuum cleaner & [vacuum\_cleaner 1001]\\
Plug out vacuum cleaner & [vacuum\_cleaner 1001]\\
\bottomrule
\end{tabular}
\end{table}

\begin{table}[h!]
\centering
\begin{tabular}{l l}
\multicolumn{2}{c}{\bf Task: Eat}\\
\midrule
Walk to kitchen & [oven 82]\\
Walk to plate & [plate 1003]\\
Find plate & [plate 1003]\\
Grab plate & [plate 1003]\\
Find food food & [food\_food 1000]\\
Grab food food & [food\_food 1000]\\
Put food food on plate & [food\_food 1000, plate 1003]\\
Walk to dining room & [dining\_room 1]\\
Walk to table & [table 63]\\
Find table & [table 63]\\
Put plate on table & [plate 1003, table 63]\\
Find chair & [chair 60]\\
Sit on chair & [chair 60]\\
Eat plate & [plate 1003]\\
Eat food food & [food\_food 1000]\\
Put food food on plate & [food\_food 2021, plate 2166]\\
\bottomrule
\end{tabular}
\end{table}

\begin{table}[h!]
\centering
\begin{tabular}{l l}
\multicolumn{2}{c}{\bf Task: Computer work}\\
\midrule
Walk to computer & [computer 379]\\
Turn to computer & [computer 379]\\
Look at computer & [computer 379]\\
Type on computer & [computer 379]\\
Switch off computer & [computer 379]\\
\bottomrule
\end{tabular}
\end{table}

\begin{table}[h!]
\centering
\begin{tabular}{l l}
\multicolumn{2}{c}{\bf Task: Clean screen}\\
\midrule
Walk to home office & [home\_office 241]\\
Walk to computer & [computer 276]\\
Find computer & [computer 276]\\
Turn to computer & [computer 276]\\
Look at computer & [computer 276]\\
Switch off computer & [computer 276]\\
Find brush & [brush 2108]\\
Turn to brush & [brush 2108]\\
Look at brush & [brush 2108]\\
Grab brush & [brush 2108]\\
Wipe computer & [computer 209]\\
Drop brush & [brush 2108]\\
Find napkin & [napkin 2060]\\
Grab napkin & [napkin 2060]\\
Wash napkin & [napkin 2060]\\
Wipe computer & [computer 276]\\
Drop napkin & [napkin 2060]\\
Find brush & [brush 2705]\\
Grab brush & [brush 2705]\\
Wipe brush & [brush 2108]\\
\bottomrule
\end{tabular}
\end{table}

\begin{table}[h!]
\centering
\begin{tabular}{l l}
\multicolumn{2}{c}{\bf Task: Look at painting}\\
\midrule
Walk to living room & [dining\_room 1]\\
Walk to painting & [drawing 139]\\
Turn to painting & [drawing 139]\\
Look at painting & [drawing 139]\\
Find painting & [drawing 139]\\
Turn to painting & [drawing 138]\\
Look at painting & [drawing 138]\\
Turn to painting & [drawing 140]\\
Look at painting & [drawing 140]\\
Turn to painting & [drawing 141]\\
Look at painting & [drawing 141]\\
Turn to painting & [drawing 142]\\
Look at painting & [drawing 142]\\
Turn to painting & [drawing 186]\\
Look at painting & [drawing 186]\\
Turn to painting & [drawing 283]\\
Look at painting & [drawing 283]\\
Turn to painting & [drawing 283]\\
\bottomrule
\end{tabular}
\end{table}

\begin{table}[h!]
\centering
\begin{tabular}{l l}
\multicolumn{2}{c}{\bf Task: Put clothes away}\\
\midrule
Find clothes dress & [clothes\_dress 2134]\\
Turn to clothes dress & [clothes\_dress 2134]\\
Point at clothes dress & [clothes\_dress 2134]\\
Wash clothes dress & [clothes\_dress 2134]\\
Turn to clothes dress & [clothes\_dress 2202]\\
Look at clothes dress & [clothes\_dress 2134]\\
Wash clothes dress & [clothes\_dress 2202]\\
Look at clothes dress & [clothes\_dress 2202]\\
Wash clothes dress & [clothes\_dress 2229]\\
Look at clothes dress & [clothes\_dress 2229]\\
Wash clothes dress & [clothes\_dress 2467]\\
Look at clothes dress & [clothes\_dress 2467]\\
Wash clothes dress & [clothes\_dress 2592]\\
Look at clothes dress & [clothes\_dress 2592]\\
\bottomrule
\end{tabular}
\end{table}

\begin{table}[h!]
\centering
\begin{tabular}{l l}
\multicolumn{2}{c}{\bf Task: Start computer}\\
\midrule
Walk to home office & [home\_office 179]\\
Walk to chair & [chair 227]\\
Pull chair & [chair 227]\\
Sit on chair & [chair 227]\\
Find computer & [computer 264]\\
Switch on computer & [computer 264]\\
Turn to computer & [computer 264]\\
Look at computer & [computer 264]\\
Turn to home office & [home\_office 179]\\
Walk to chair & [chair 133]\\
Walk to chair & [chair 134]\\
Pull chair & [chair 134]\\
Sit on chair & [chair 134]\\
Find computer & [computer 320]\\
Switch on computer & [computer 320]\\
Turn to computer & [computer 320]\\
Look at computer & [computer 320]\\
Turn to home office & [home\_office 179]\\
\bottomrule
\end{tabular}
\end{table}

\begin{table}[h!]
\centering
\begin{tabular}{l l}
\multicolumn{2}{c}{\bf Task: Compute}\\
\midrule
Walk to home office & [home\_office 1]\\
Walk to chair & [chair 17]\\
Find chair & [chair 17]\\
Sit on chair & [chair 17]\\
Find computer & [computer 98]\\
Switch on computer & [computer 98]\\
Find notebook & [laptop 2077]\\
Grab notebook & [laptop 2077]\\
\bottomrule
\end{tabular}
\end{table}

\begin{table}[h!]
\centering
\begin{tabular}{l l}
\multicolumn{2}{c}{\bf Task: Straighten pictures on wall}\\
\midrule
Walk to living room & [dining\_room 41]\\
Walk to wall & [wall 68]\\
Find picture & [picture 2114]\\
Turn to picture & [picture 2114]\\
Look at picture & [picture 2114]\\
Grab picture & [picture 2114]\\
Walk to wall & [wall 2]\\
Put picture on wall & [picture 2114, wall 2]\\
Find picture & [picture 2331]\\
Turn to picture & [picture 2331]\\
Look at picture & [picture 2331]\\
Grab picture & [picture 2331]\\
Walk to wall & [wall 3]\\
Put picture on wall & [picture 2331, wall 3]\\
Find picture & [picture 2402]\\
Grab picture & [picture 2402]\\
Walk to wall & [wall 4]\\
Put picture on wall & [picture 2402, wall 4]\\
Find picture & [picture 2417]\\
Grab picture & [picture 2417]\\
\bottomrule
\end{tabular}
\end{table}

\begin{table}[h!]
\centering
\begin{tabular}{l l}
\multicolumn{2}{c}{\bf Task: Work}\\
\midrule
Walk to desk & [desk 104]\\
Walk to chair & [chair 103]\\
Find chair & [chair 103]\\
Sit on chair & [chair 103]\\
Find laptop & [laptop 2120]\\
Open laptop & [laptop 2120]\\
\bottomrule
\end{tabular}
\end{table}

\begin{table}[h!]
\centering
\begin{tabular}{l l}
\multicolumn{2}{c}{\bf Task: Put away keys}\\
\midrule
Walk to entrance hall & [dining\_room 1]\\
Walk to purse & [purse 2067]\\
Find purse & [purse 2067]\\
Open purse & [purse 2067]\\
Find keys & [keys 1000]\\
Grab keys & [keys 1000]\\
Close purse & [purse 2067]\\
\bottomrule
\end{tabular}
\end{table}

\begin{table}[h!]
\centering
\begin{tabular}{l l}
\multicolumn{2}{c}{\bf Task: Spread table with appropriate supplies}\\
\midrule
Walk to table & [table 123]\\
Put food food in table & [food\_food 2006, table 123]\\
Put food food in bowl & [food\_food 2006, bowl 2284]\\
\bottomrule
\end{tabular}
\end{table}

\begin{table}[h!]
\centering
\begin{tabular}{l l}
\multicolumn{2}{c}{\bf Task: Send  email}\\
\midrule
Walk to chair & [chair 103]\\
Sit on chair & [chair 103]\\
Find laptop & [laptop 2053]\\
Switch on laptop & [laptop 2053]\\
Turn to laptop & [laptop 2053]\\
Look at laptop & [laptop 2053]\\
Find mail & [mail 1000]\\
Grab mail & [mail 1000]\\
Read mail & [mail 1000]\\
Find keyboard & [keyboard 168]\\
Type on keyboard & [keyboard 168]\\
Switch off laptop & [laptop 2053]\\
\bottomrule
\end{tabular}
\end{table}

\begin{table}[h!]
\centering
\begin{tabular}{l l}
\multicolumn{2}{c}{\bf Task: Put in chair}\\
\midrule
Walk to dining room & dining\_room (100]\\
Walk to chair & chair (126]\\
Put chair in table & chair 126, table 128]\\
Sit on chair & chair (126]\\
Put chair on table & chair 126, table 128]\\
Sit on chair & chair (127]\\
Put back chair & chair (127]\\
Sit on chair & chair (124]\\
Put back chair & chair (124]\\
Sit on chair & chair (125]\\
Put back chair & chair (125]\\
Sit on chair & chair (17]\\
Put back chair & chair (17]\\
Sit on chair & chair (240]\\
Put back chair & chair (240]\\
Sit on chair & chair (240]\\
Put back chair & chair (126]\\
Sit on chair & chair (126]\\
Put back chair & chair (126]\\
Sit on chair & chair (2002]\\
\bottomrule
\end{tabular}
\end{table}

\begin{table}[h!]
\centering
\begin{tabular}{l l}
\multicolumn{2}{c}{\bf Task: Open curtains}\\
\midrule
Walk to home office & [home\_office 241]\\
Walk to curtain & [curtain 289]\\
Find curtain & [curtain 289]\\
Pull curtain & [curtain 289]\\
Open curtain & [curtain 289]\\
Walk to curtain & [curtain 290]\\
Find curtain & [curtain 290]\\
Pull curtain & [curtain 290]\\
Open curtain & [curtain 290]\\
Walk to home office & [home\_office 241]\\
Walk to curtain & [curtain 291]\\
Find curtain & [curtain 291]\\
Pull curtain & [curtain 291]\\
Open curtain & [curtain 291]\\
Walk to curtain & [curtain 22]\\
Find curtain & [curtain 22]\\
Pull curtain & [curtain 22]\\
Open curtain & [curtain 22]\\
Walk to curtain & [curtain 206]\\
Find curtain & [curtain 206]\\
\bottomrule
\end{tabular}
\end{table}

\end{document}